\documentclass[11pt]{article}
\usepackage{algorithm}
\usepackage{algorithmicx}
\usepackage{algpseudocode}
\usepackage{amsmath}
\usepackage{booktabs}
\usepackage{multirow}
\usepackage[]{acl}
\usepackage{placeins}  
\usepackage{array}  
\usepackage{caption}
\usepackage{times}
\usepackage{latexsym}
\usepackage{comment}
\usepackage[T1]{fontenc}

\usepackage[utf8]{inputenc}
\usepackage{microtype}

\usepackage{inconsolata}

\usepackage{graphicx}

\title{%
  \begin{tabular}{@{}l@{}}
    \raisebox{-0.2\height}{\includegraphics[height=1.8em]{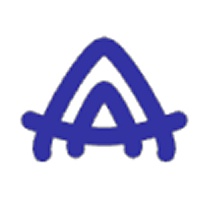}} AI2Agent: An End-to-End Framework \\[1mm]
    for Deploying AI Projects as Autonomous Agents
  \end{tabular}%
}
\author{
    \textbf{Jiaxiang Chen}\textsuperscript{1,2}, 
    \textbf{Jingwei Shi}\textsuperscript{1}, 
    \textbf{Lei Gan}\textsuperscript{1}, 
    \textbf{Jiale Zhang}\textsuperscript{1}, 
    \textbf{Qingyu Zhang}\textsuperscript{1}, \\
    \textbf{Dongqian Zhang}\textsuperscript{1}, 
    \textbf{Xin Pang}\textsuperscript{1*}, 
    \textbf{Zhucong Li}\textsuperscript{1,2*}, 
    \textbf{Yinghui Xu}\textsuperscript{2}
    \\
    $^1$ Continue-AI \\
    $^2$ Artificial Intelligence Innovation and Incubation Institute, \\
    Fudan University, Shanghai, China \\
    pxavdpro@gmail.com, 
    \{jiaxiangchen23,zcli22\}@m.fudan.edu.cn\\
    \{xuyinghui\}@fudan.edu.cn
}

\begin{document}
\maketitle
{
\renewcommand{\thefootnote}{\fnsymbol{footnote}}
\footnotetext[1]{Corresponding author.}
}
\begin{abstract}
As AI technology advances, it is driving innovation across industries, increasing the demand for scalable AI project deployment. However, deployment remains a critical challenge due to complex environment configurations, dependency conflicts, cross-platform adaptation, and debugging difficulties, which hinder automation and adoption.
This paper introduces AI2Agent, an end-to-end framework that automates AI project deployment through guideline-driven execution, self-adaptive debugging, and case \& solution accumulation. AI2Agent dynamically analyzes deployment challenges, learns from past cases, and iteratively refines its approach, significantly reducing human intervention.
To evaluate its effectiveness, we conducted experiments on 30 AI deployment cases, covering TTS, text-to-image generation, image editing, and other AI applications. Results show that AI2Agent significantly reduces deployment time and improves success rates. The code\footnote{\url{https://github.com/continue-ai-company/AI2Agent}} and demo video\footnote{\url{https://youtu.be/seRTYtwgLrk}} are now publicly accessible.

\end{abstract}

\section{Introduction}



\begin{figure}[thb]
	\centering
	\includegraphics[scale=0.3]{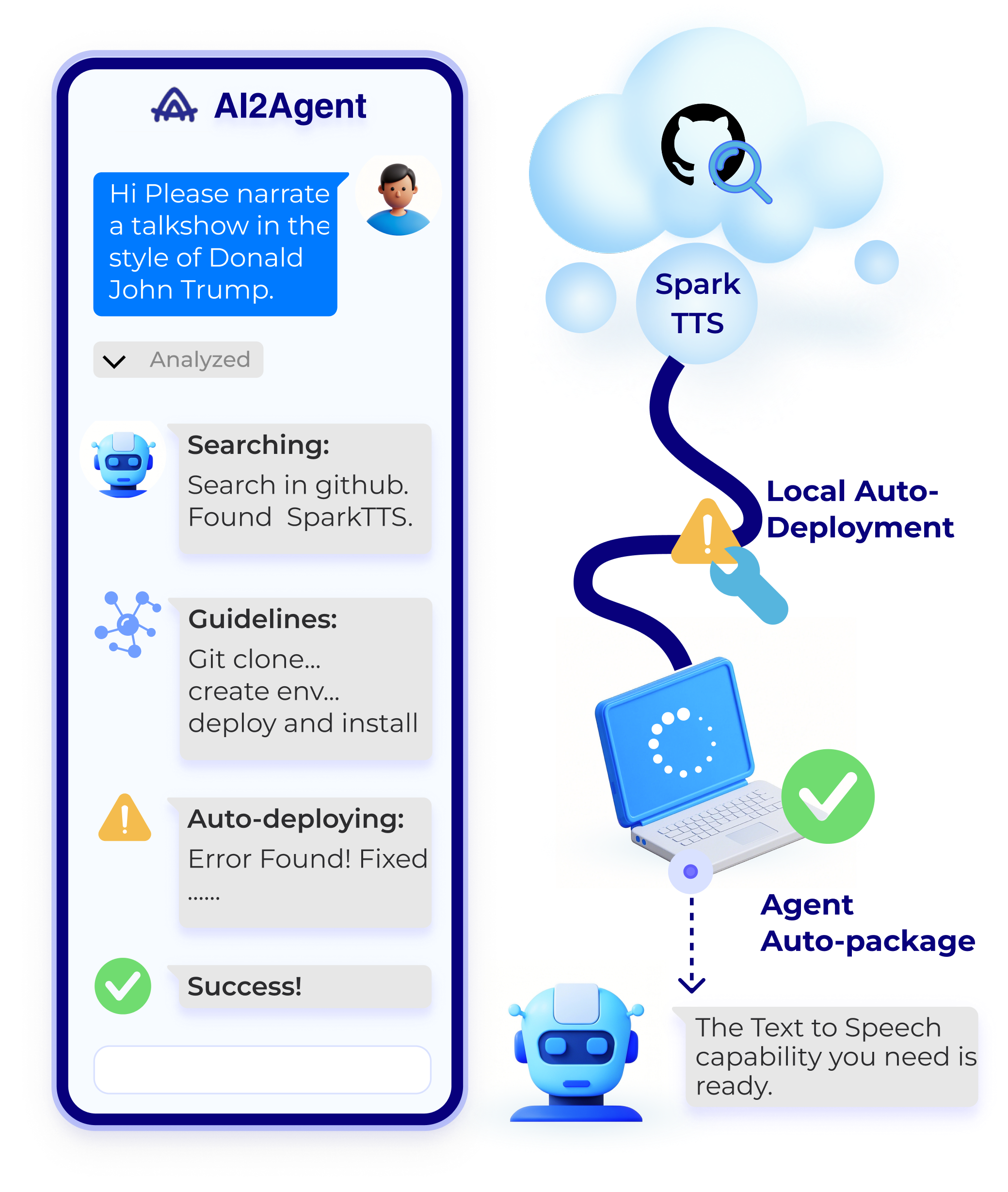}
	\caption{
		\textbf{Left:} The AI2Agent user interface, illustrating the automated workflow where a user request (e.g., generating a talk show in a specific style) initiates a structured execution process. This includes searching for a suitable project, following the predefined guidelines for execution, auto-deployment and debug to ensure success.  
		\textbf{Right:} A conceptual visualization of local auto-deployment and Agent auto-package, demonstrating how AI2Agent transforms text-to-speech(TTS) functionality into a fully autonomous agent.
	}
	\label{fig:value}
\end{figure}

AI is revolutionizing industries, from autonomous driving to healthcare and finance. However, to fully realize its potential, AI must be effectively deployed into diverse environments. Deployment remains a major bottleneck due to complex environment configurations, dependency conflicts, and cross-platform adaptation issues, which hinder scalability and adoption.By automating deployment and debugging, AI can be more efficiently integrated across diverse domains, reducing engineering barriers and accelerating innovation at scale.

Automated deployments in industry predominantly follow DevOps paradigm~\cite{bass2015devops,ebert2016devops}, where execution environments, dependencies, and orchestration rules are defined using static configuration files (e.g., YAML). In this approach, developers manually configure environments, specifying software packages, version constraints, and computational resource allocations. CI/CD pipelines are then employed to automate building, testing, and deployment. To enhance automation, tools like AutoDevOps~\cite{AutoDevops} utilize predefined templates that streamline the configuration process. 

However, these methods have significant limitations. First, they lack flexibility, as they cannot adjust deployment strategies based on runtime conditions. This often leads to manual fixes for dependency conflicts, environment mismatches, and hardware compatibility issues. Second, they fail to retain and reuse past solutions, meaning each deployment starts from scratch. Developers must repeatedly troubleshoot the same problems, making the process inefficient and time-consuming. Third, they do not support scalable AI workflows, as they focus on deploying isolated models rather than integrating multiple AI components. Real-world AI applications often require chaining models or services together, but without standardized interfaces, these tools cannot effectively automate such tasks. As a result, deployment remains fragmented, difficult to scale, and heavily dependent on manual intervention.

To overcome these limitations, we introduce AI2Agent, an end-to-end framework that transforms AI projects into Autonomous Agents, enabling adaptive and reusable deployments. Unlike static DevOps paradigm, AI2Agent leverages autonomous execution, real-time debug, and experience accumulation to stabilize deployment processes.It consists of three components: Guideline-driven Execution, Self-adaptive Debug, and Case \& Solution Accumulation. Guideline-driven Execution ensures repeatable and structured deployments by following predefined guideline steps, such as searching for dependencies and executing system commands. Unlike static scripts, AI2Agent adapts dynamically to various environments. Self-adaptive Debug enhances deployment reliability by adjusting strategies based on real-time feedback, autonomously troubleshooting issues like search online or query the knowledge repository. Finally, Case \& Solution Accumulation utilizes a Knowledge Repository and RAG~\cite{graphrag,lightrag,ragflow}
to store past experiences, continuously refining deployment strategies and reducing errors, leading to more efficient deployments over time.

To validate the capabilities of AI2Agent, we conducted a case study on the automated deployment of multiple AI applications, including TTS~\cite{sparktts}, text-to-image generation~\cite{dall-e}, and image editing~\cite{instructpix2pix}. As shown in Figure~\ref{fig:value}, AI2Agent autonomously searches for and executes deployments, following predefined guidelines, and successfully packages them as Agents through self-adaptive debug. Test results show that AI2Agent significantly shortened deployment time, improved success rates, and reduced errors, leading to faster and more reliable deployments. This demonstrates its potential in enabling standardized, modular, and reusable AI deployments, paving the way for a more autonomous and interoperable AI ecosystem.

The key contributions of this paper are as follows:
\begin{itemize}
    \item We introduce \textbf{AI2Agent}, an end-to-end framework that automates AI project deployment by transforming them into autonomous Agents. It provides a standardized Agent interface, enabling modular management, seamless execution, and improved reusability, fostering a more interoperable AI ecosystem.
    \item Our approach comprises \textbf{Guideline-driven Execution, Self-adaptive Debug, and Case \& Solution Accumulation}, forming a structured yet flexible framework. It follows predefined guidelines while dynamically adapting to deployment environments and continuously improving through accumulated experience.
    \item We evaluate AI2Agent across \textbf{30 AI projects}, covering areas such as TTS, text-to-image generation, and image editing. Experimental results show \textbf{significant reductions in deployment time and error rates}, demonstrating the effectiveness and reliability of our approach in streamlining AI deployment.
\end{itemize}

\section{Related Work}

\subsection{DevOps Paradigm}
Automated deployments in industry predominantly follow the DevOps paradigm~\cite{bass2015devops,ebert2016devops}, where execution environments, dependencies, and orchestration rules are defined using static configuration files (e.g., YAML). Developers manually specify dependencies, version constraints, and resource allocations, while CI/CD pipelines automate the build, test, and deployment processes. While this improves reproducibility, adapting to new environments, debugging failures, and handling model updates in AI projects still require significant manual effort.

To enhance automation, AutoDevOps~\cite{AutoDevops} offers predefined templates that simplify configuration and deployment. While effective for standardized workflows, its static nature limits adaptability, making it challenging to handle the complexity and variability of AI deployments. AI projects often require integrating multiple models, dynamically managing resource constraints, and resolving intricate dependency conflicts—tasks that demand greater flexibility. Although some approaches~\cite{battina2019intelligentdevops,karamitsos2020applyingdevops,bou2024enhancingdevops,enemosah2025enhancingdevops} incorporate machine learning or LLM to improve automation, they still lack the autonomy and intelligence needed to independently adapt to evolving deployment requirements.

\begin{figure*}[h]
    \centering
    \includegraphics[scale=0.45]{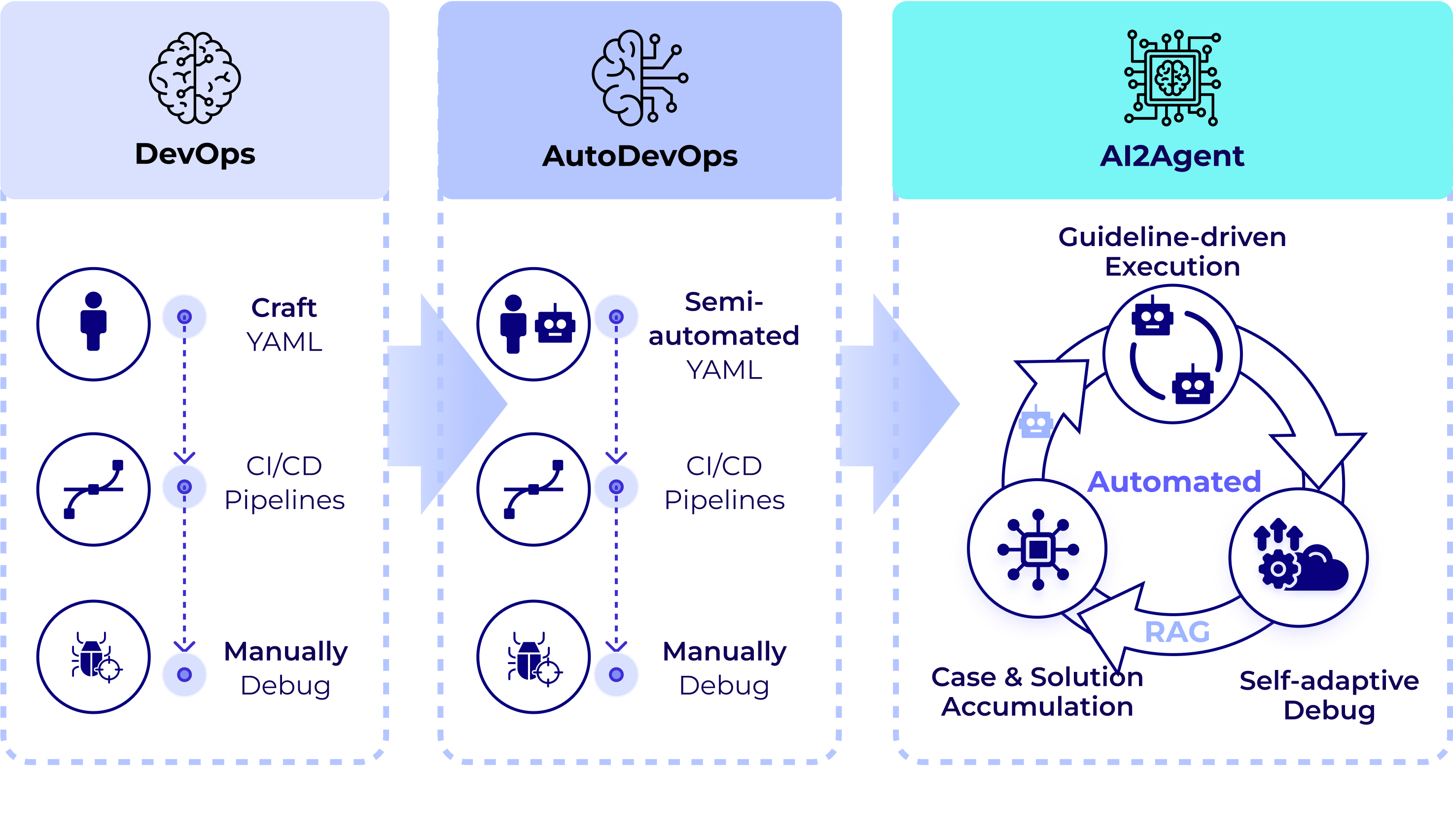}
    \caption{Comparison of Paradigms. DevOps relies on manual YAML configuration and CI/CD workflows with manual debug. AutoDevOps offers semi-automated configuration but still requires human intervention. AI2Agent achieves end-to-end automated performance, including guideline-driven execution, self-adaptive debug, and case \& solution accumulation.}
    \label{fig:cmp_diagm}
\end{figure*}

As shown in Figure \ref{fig:cmp_diagm}, AI2Agent addresses these challenges through three key components: Guideline-driven Execution, Self-adaptive Debug, and Case \& Solution Accumulation. By following structured deployment guidelines, autonomously identifying and fixing errors, and continuously refining deployment strategies based on past cases, AI2Agent significantly reduces manual intervention. This enables more flexible, efficient, and scalable AI deployments across diverse environments.

\subsection{LLM-Based Agents}

Large Language Models (LLMs)~\cite{GPTs} excel in reasoning and decision-making but struggle with executing actions and using external tools. AI agents address this by integrating LLMs with structured tool use, enhancing automation and adaptability.

ReAct~\cite{react} enables agents to iteratively reason, act, and observe but lacks mechanisms to draw on past experiences, thus limiting long-term planning. AutoGPT~\cite{autogpt} improves autonomy through multi-step planning yet struggles with execution reliability in real-world scenarios. MetaGPT~\cite{metagpt} enhances multi-agent collaboration but remains constrained by workflow adaptability. AutoGen~\cite{autogen} introduces agent collaboration but faces issues with scalability and consistency.

Existing approaches either focus on reasoning without leveraging past experiences or improve execution with tools but lack adaptive workflow management. AI2Agent addresses this gap by integrating experience-driven learning, structured tool use, and dynamic workflow orchestration, ensuring efficient and adaptable AI deployment.

\section{Method}

\subsection{Overview of the AI2Agent Framework}
As illustrated in Algorithm~\ref{alg:general_deploy}, AI2Agent is an end-to-end framework designed to transform AI projects into autonomous agents, enhancing both automation and reusability in deployment. Unlike traditional DevOps and AutoDevOps, which rely on manual configurations or static templates, AI2Agent integrates intelligent reasoning, automated debugging, and iterative experience accumulation to enable dynamic and adaptive deployment. The framework consists of three core modules: \textbf{Guideline-Driven Execution}, which standardizes deployment steps; \textbf{Self-Adaptive Debug}, which resolves issues through automated analysis; and \textbf{Case \& Solution Accumulation}, which continuously refines deployment strategies based on past experiences.

\subsection{Guideline-Driven Execution}
\begin{figure*}[tbh]
	\centering
	\includegraphics[scale=0.4]{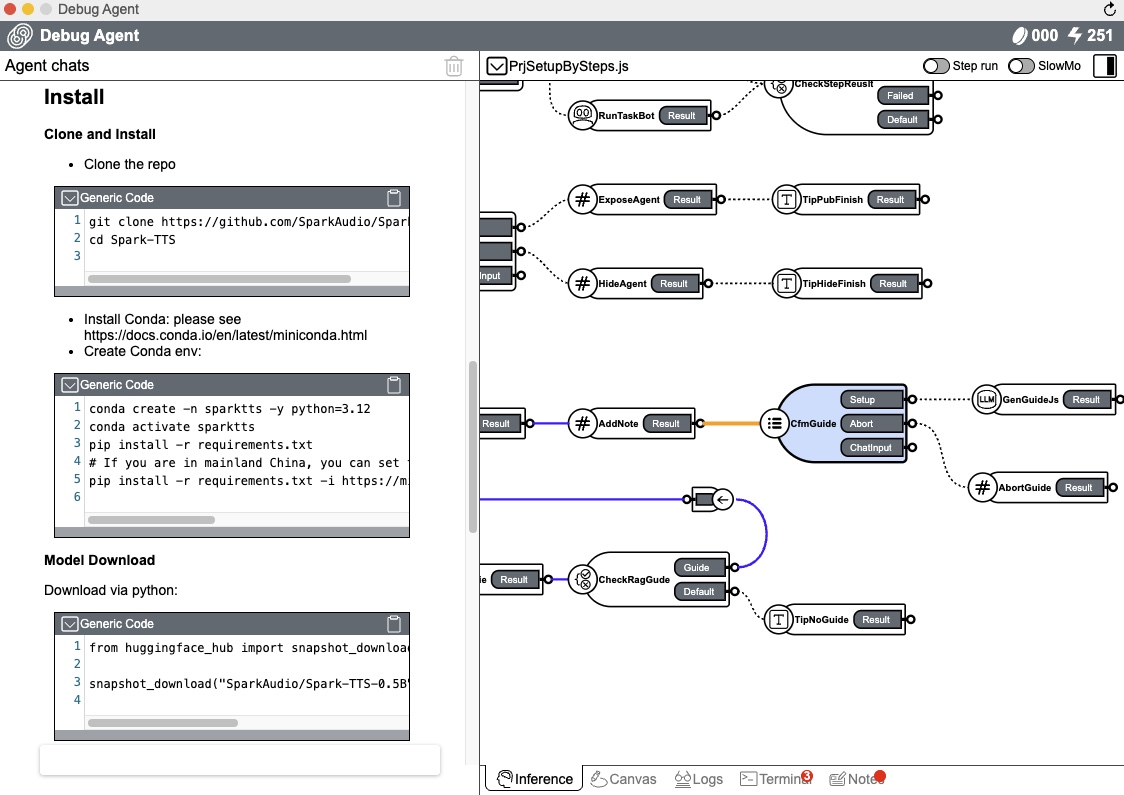}
	\caption{Screenshot of our local auto-deployment process. \textbf{Left:} Execution following predefined guidelines to ensure structured and reliable deployment. \textbf{Right:} The inference and planning interface for dynamically adapting to deployment conditions.}
	\label{fig:guideline}
\end{figure*}

AI2Agent adopts a guideline-driven execution strategy to ensure a structured, efficient, and reliable deployment process. Instead of relying solely on autonomous adjustments, it follows predefined, validated workflows that incorporate best practices and accumulated expertise, as illustrated in Figure~\ref{fig:guideline}.

By adhering to these step-by-step guidelines, AI2Agent minimizes uncertainty and maintains consistency across deployments. In complex or ambiguous scenarios, it proactively references its Knowledge Repository to retrieve relevant cases and proven solutions, thereby reducing unnecessary trial-and-error debugging.

This structured approach not only improves deployment stability and efficiency but also mitigates potential failures by grounding the execution process in historical insights and validated experience.

\subsection{Self-Adaptive Debug}
To address the dynamic nature of deployment environments, AI2Agent integrates a self-adaptive debugging mechanism that refines execution strategies based on real-time feedback. This mechanism consists of three key components:

\paragraph{Step-by-Step Execution} AI2Agent dynamically adjusts its execution flow in response to real-time environmental feedback. During deployment, it continuously monitors system parameters such as computational resources, dependency versions, and runtime errors. Based on this data, AI2Agent adjusts execution parameters to enhance performance and maintain system stability.

\paragraph{Environment-Aware Debug} When deployment failures occur, AI2Agent automatically diagnoses issues by analyzing execution logs and system constraints. It identifies failure patterns, refines its execution strategy, and applies corrective actions to enhance robustness. If necessary, AI2Agent can perform an online search to gather additional information or solutions, further improving its problem-solving capabilities. This proactive approach reduces disruptions and ensures smoother execution.

\paragraph{Knowledge-Guided Refinement} To improve debugging efficiency, AI2Agent queries its Knowledge Repository for relevant failure cases and proven solutions. By leveraging historical insights, AI2Agent accelerates problem resolution and refines debugging techniques, increasing deployment success rates while minimizing human intervention.

\subsection{Case \& Solution Accumulation}
AI2Agent continuously refines its deployment strategies by accumulating experience. The Knowledge Repository serves as a structured database that records successful deployment cases, failure resolutions, and improvement strategies. This module enables two key processes:

\paragraph{Retrieval of Deployment Insights} During deployment, AI2Agent retrieves relevant historical cases using Retrieval-Augmented Generation (RAG). These insights help in selecting optimal configurations, dependency management strategies, and execution plans tailored to the current task.

\paragraph{Continuous Learning and Refinement} Every deployment contributes new insights to the repository. AI2Agent analyzes both successful and failed deployments, extracting lessons from error logs and corrective actions. These insights are integrated into future executions, ensuring a continuously evolving and increasingly autonomous deployment framework.

\begin{algorithm}
\caption{AI2Agent: Automated AI Deployment}
\label{alg:general_deploy}
\begin{algorithmic}[1]

\Require Repository $R$, Execution Environment $\mathcal{E}$ 
\Return Successfully deployed AI Agent $\mathcal{A}$

\State \textbf{\textcolor{teal}{// Guideline-driven Execution}}  
\State ${G} \gets f_{\text{load}}({R})$  
\State $S \gets \emptyset$ // Store execution results

\For{each step ${G}_t \in {G}$}
    \State \textbf{// Execute step} 
    \State $S_t \gets f_{\text{exec}}(\mathcal{E}, {G}_t)$  
    \State $S \gets S \cup \{S_t\}$  
   \State \textbf{\textcolor{teal}{// Self-adaptive Debug}}  
    \While{$f_{\text{status}}(S_t) = 0$} 
         \State ${G}_t' \gets f_{\text{search}}(S_t,R)$   {//search solutions}
         \State $S_t' \gets f_{\text{exec}}(\mathcal{E}, {G}_t')$  
         \State $S \gets S \cup \{S_t'\}$
     \State \textbf{\textcolor{teal}{// Case \& Solution Accumulation}}
     \State ${R} \gets f_{\text{merge}}({G}_t',S_t')$
    \EndWhile 
\EndFor 

\State \textbf{\textcolor{teal}{// Auto-Deployed AI Agent}}  
\State $\mathcal{A} \gets f_{\text{auto-deploy}}(S)$  
\State \textbf{return} $\mathcal{A}$  

\end{algorithmic}
\end{algorithm}

\section{Case Study}

\begin{figure}[tb]
	\centering
	\includegraphics[scale=0.4]{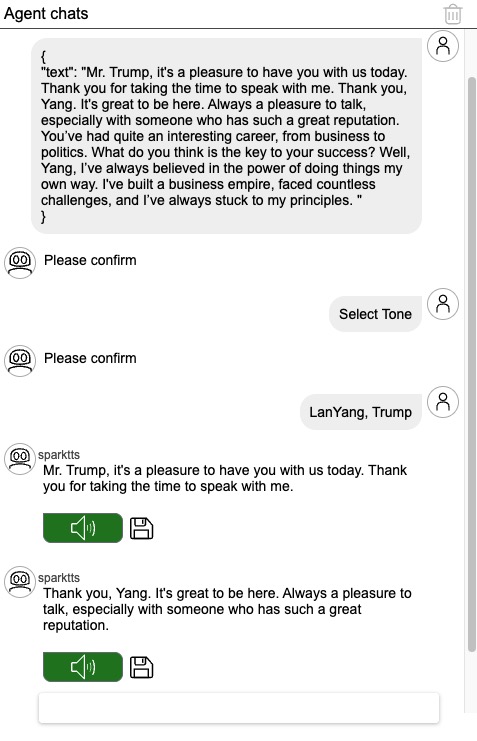}
	\caption{Screenshot of the user interface after auto-deployment.}
	\label{fig:gen}
\end{figure}


\begin{figure}[tb]
    \centering
    \includegraphics[scale=0.4]{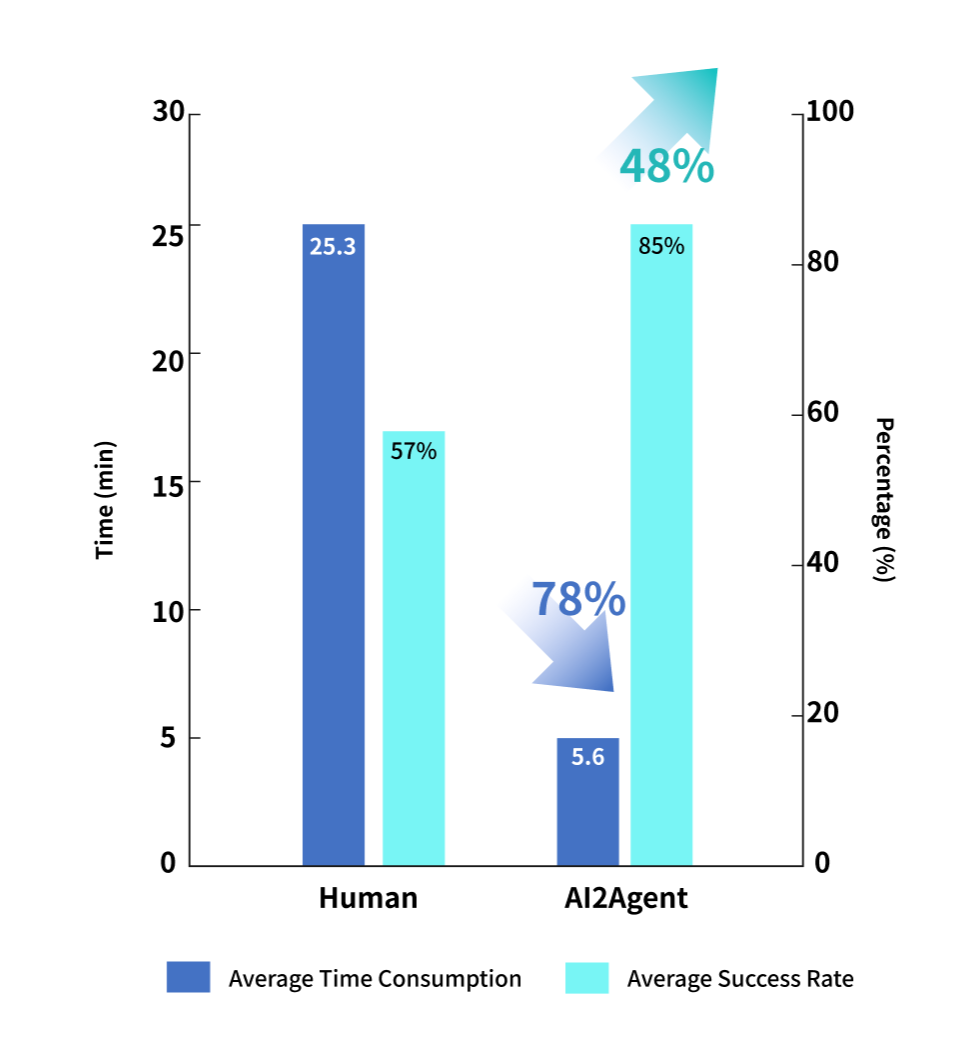}
    \caption{
        Comparison of manual vs. AI2Agent deployment:  
        (1) \textbf{Time Consumption}: AI2Agent reduces deployment time by 78\%.  
        (2) \textbf{Success Rate}: AI2Agent improves success rate by 48\%.  
    }
    \label{fig:score}
\end{figure}

To evaluate the effectiveness of AI2Agent in automating AI project deployments, we conducted a comprehensive study across 30 AI applications, including text-to-speech (TTS), text-to-image generation, and image editing. Figure~\ref{fig:gen} showcases a sample user interface from a successfully deployed project. This section further details details our deployment environment, evaluation setup, and a demo case of Spark-TTS.

\subsection{Deployment Environment}
We conducted all experiments using our self-developed \textbf{AI2Apps IDE~\cite{ai2apps}}, which is based on a web container framework. AI2Apps IDE is available in both web-based and locally deployed versions. To ensure security and full control over execution, all cases in this study were conducted using the locally deployed version. This approach guarantees that AI project installations and configurations are fully automated within a sandboxed environment, mitigating potential security risks associated with external dependencies and permissions.

\subsection{Evaluation Setup}
We assessed AI2Agent's performance across 30 AI applications spanning multiple domains, including text-to-speech (TTS), text-to-image generation, and image editing. To establish a fair comparison, we recruited participants with deep learning experience and provided them with guideline documents for manual deployment. Each participant independently attempted to deploy the applications without automation assistance.

To quantify AI2Agent’s effectiveness, we evaluated two key metrics:
\textbf{Deployment Time Reduction}: The average time required for installation and configuration, excluding model download time, as it is influenced by network speed.
\textbf{Success Rate Improvement}: The percentage of successful deployments completed without additional troubleshooting.

As shown in Figure~\ref{fig:score}, AI2Agent achieved a \textbf{78\%} reduction in average deployment time and a \textbf{48\%} increase in overall success rate, demonstrating its ability to streamline and enhance AI deployment efficiency.

\subsection{Demonstration: Deploying Spark-TTS}
To showcase AI2Agent’s automation capabilities, we present a case study on deploying \textbf{Spark-TTS}, a powerful text-to-speech (TTS) system. AI2Agent autonomously manages the entire setup process, including dependency management, model configuration, and environment preparation, ensuring a streamlined and error-free deployment. As illustrated in the user interface shown in Figure~\ref{fig:gen} and video\href{https://github.com/Avdpro/ai2apps}, AI2Agent significantly reduces manual effort while maintaining a high success rate. The final speech output demonstrates excellent pronunciation accuracy, fluency, and overall quality, further validating the effectiveness of the framework.

\section{Conclusion}
Deploying AI projects is often hindered by complex environment configurations, dependency conflicts, and debugging challenges, limiting automation and scalability. While DevOps and AutoDevOps improve automation through YAML configurations and CI/CD pipelines, they still require manual intervention and lack adaptability in dynamic environments.
This paper introduced AI2Agent, an end-to-end framework for automating AI deployment. By integrating guideline-driven execution, self-adaptive debugging, and case \& solution accumulation, AI2Agent minimizes manual effort and dynamically adapts to deployment challenges. Over time, it learns from past deployments, enhancing efficiency and success rates.
Experiments on 30 AI applications showed that AI2Agent reduces deployment time by 78\% and increases success rates by 48\%, demonstrating its potential to streamline AI deployment. This framework provides a scalable, automated solution for AI adoption across industries.

\section*{Limitations}
AI2Agent enhances automation and adaptability in AI deployment, yet there is always room for further refinement. As AI projects grow increasingly complex, we look forward to exploring new ways to make deployments even more seamless and intelligent. We welcome researchers and practitioners to contribute to AI2Agent’s evolution, helping to expand its capabilities and applications.

\section*{Ethics Statement}
(1)This work is the authors’ original research and has not been previously published elsewhere.
(2)The paper is not under consideration for publication in any other venue.
The research is conducted with integrity, ensuring truthful and complete reporting of methods, findings, and limitations.
(3)AI2Agent does not involve the collection of personally identifiable information or sensitive user data. Any case study participants were volunteers who provided informed consent before participation, and all identifiers used in experiments were anonymized.
(4)AI2Agent is designed to enhance the deployment of AI applications, promoting efficiency while ensuring responsible AI practices.
(5)Our work does not involve training or fine-tuning large language models (LLMs); it strictly utilizes publicly available APIs permitted for research purposes, ensuring compliance with ethical and legal standards.

\bibliography{references}

\end{document}